\documentclass[]{article}
\usepackage{proceed2e}
\usepackage{times}

\usepackage{hyperref}
\usepackage[utf8]{inputenc}
\DeclareUnicodeCharacter{00A0}{ }
\usepackage[authoryear]{natbib}
\usepackage{amsmath}
\usepackage{amsthm}
\usepackage{amssymb}
\usepackage{xcolor}
\usepackage{pst-plot}
\usepackage{floatrow}
\usepackage[font=small,labelfont=bf]{caption}
\usepackage{pstricks}
\usepackage{enumerate}
\usepackage{colortbl}
\usepackage{multirow}
\usepackage{tikz}
\usepackage{xr}
\usepackage{placeins}
\usepackage{floatrow}
\usepackage{arydshln}
\usepackage[caption=false]{subfig}
\newtheorem{theorem}{Theorem}

\newtheorem{definition}{Definition}
\newtheorem{example}{Example}

\newcommand{\g}{\cellcolor{gray!20}}
\newcolumntype{L}[1]{>{\raggedright\let\newline\\\arraybackslash\hspace{0pt}}m{#1}}
\newcolumntype{C}[1]{>{\centering\let\newline\\\arraybackslash\hspace{0pt}}m{#1}}
\newcolumntype{R}[1]{>{\raggedleft\let\newline\\\arraybackslash\hspace{0pt}}m{#1}}

\title{Context-dependent feature analysis with random forests}
\author{Antonio Sutera$^1$ \And Gilles Louppe$^2$ \\ $^1$ Dept. of EE \& CS, University of Liège, Belgium \And Vân Anh Huynh-Thu$^1$ \And  Louis Wehenkel$^1$ \\
 $^2$ New York University, USA \And Pierre Geurts$^1$}

\begin{document}

%

%



\maketitle
\begin{abstract}

In many cases, feature selection is often more complicated than
identifying a single subset of input variables that would together
explain the output. There may be interactions that depend on
contextual information, i.e., variables that reveal to be relevant
only in some specific circumstances. In this setting, the contribution
of this paper is to extend the random forest variable importances
framework in order (i) to identify variables whose relevance is
context-dependent and (ii) to characterize as precisely as possible
the effect of contextual information on these variables.
The usage and the relevance of our framework for highlighting context-dependent variables
is illustrated on both artificial and real datasets.

\end{abstract}


\section{\textsc{Motivation}}
\label{sec:motivation}

Supervised learning finds applications in many domains such as
medicine, economics, computer vision, or bioinformatics. Given a
sample of observations of several inputs and one output variable, the
goal of supervised learning is to learn a model for predicting the
value of the output variable given any values of the input
variables. Another common side objective of supervised learning is to
bring as much insight as possible about the relationship between the
inputs and the output variable. One of the simplest ways to gain such
insight is through the use of feature selection or ranking methods
that identify the input variables that are the most decisive or
relevant for predicting the output, either alone or in combination
with other variables. Among feature selection/ranking methods, one
finds variable importance scores derived from random forest models
that stand out from the literature mainly because of their
multivariate and non parametric nature and their reasonable
computational cost.  Although very useful, feature selection/ranking
methods however only provide very limited information about the often
very complex input-output relationships that can be modeled by
supervised learning methods. There is thus a high interest in
designing new techniques to extract more complete information about
input-output relationships than a single global feature subset or
feature ranking.

In this paper, we specifically address the problem of the
identification of the input variables whose relevance or irrelevance
for predicting the output only holds in specific circumstances, where
these circumstances are assumed to be encoded by a specific context
variable. This context variable can be for example a standard input
variable, in which case, the goal of contextual analyses is to better
understand how this variable interacts with the other inputs for
predicting the output. The context can also be an external variable
that does not belong to the original inputs but that may nevertheless
affect their relevance with respect to the output. Practical
applications of such contextual analyses are numerous. E.g., one may be interested in finding variables that are both
relevant and independent of the context, as in medical
studies \citep[see, e.g.,][]{geissler2000risk}, where one is often interested
in finding risk factors that are as independent as possible of
external factors, such as the sex of the patients, their origins or
their data cohort. By contrast, in some other
cases, one may be interested in finding variables that are relevant
but dependent in some way on the context. For example, in systems
biology, differential analysis~\citep{ideker2012differential} aims at
discovering genes or factors that are relevant only in some specific
conditions, tissues, species or environments.

Our contribution in this paper is two-fold. First, starting from
common definitions of feature relevance, we propose
a formal definition of context-dependent variables and provide a
complete characterization of these variables depending on how their
relevance is affected by the context variable. Second, we extend the
random forest variable importances framework in order to identify and
characterize variables whose relevance is context-dependent or
context-independent. Building on existing theoretical results for
standard importance scores, we propose asymptotic guarantees for the
resulting new measures.

The paper is structured as follows. In
Section~\ref{sec:contextual-relevance}, we first lay out our formal
framework defining context-dependent variables and describing how the
context may change their relevance.  We describe in
Section~\ref{sec:contextual-relevance-trees} how random forest
variable importances can be used for identifying context-dependent
variables and how the effect of contextual information on these
variables can be highlighted. Our results are then illustrated in
Section~\ref{sec:experiments} on representative problems. Finally,
conclusions and directions of future works are discussed in
Section~\ref{sec:conclusions}.


\section{\textsc{Context-dependent feature selection and characterization}}
\label{sec:contextual-relevance}

\paragraph{Context-dependence.}

Let us consider a set $V = \{X_1,\dots,X_p\}$ of $p$ input variables and an
output $Y$ and let us denote by $V^{-m}$ the set $V\setminus \{X_m\}$. All
input and output variables are assumed to be categorical, not necessarily
binary\footnote{Non categorical outputs are discussed in
  Section~\ref{sec:generalisation}.}. The standard definitions of relevant,
irrelevant, and marginally relevant variables based on their mutual information
$I$ are as follows~\citep{kohavi1997wrappers,guyon2003introduction}:
\begin{itemize}
\item A variable $X_m$ is {\it relevant to $Y$ with respect to $V$}
  iff there exists a subset $B\subseteq V^{-m}$ (possibly empty) such
  that $I(Y;X_m|B)>0$.
\item A variable $X_m$ is {\it irrelevant to $Y$ with respect to $V$} iff, for all $B\subseteq V^{-m}$, $I(Y;X_m|B)=0$.
\item A variable is {\it marginally relevant to $Y$} iff $I(Y;X_m)>0$.
\end{itemize}

Let us now assume the existence of an additional (observed) context variable
$X_c\notin V$, also assumed to be categorical.
Inspired by the notion of relevant and irrelevant variables, we propose to
define context-dependent and context-independent variables as follows:
\begin{definition}\label{defcontextdependent}
A variable $X_m \in V$ is {\em context-dependent to $Y$ with respect to $X_c$}
iff there exists a subset $B \subseteq V^{-m}$ and some values $x_c$ and $b$
such that\footnote{In this definition and all definitions that follow, we
  assume that the events on which we are conditioning have a non-zero
  probability and that if such event does not exist then the condition of the
  definition is not satisfied.}:
\begin{equation}\label{eqn:context-dependent}
 I(Y;X_m|B=b,X_c=x_c) \ne I(Y;X_m|B=b).
\end{equation}
\end{definition}
\begin{definition}\label{defcontextindependent}
A variable $X_m \in V$ is {\em context-independent to $Y$ with respect to $X_c$} iff for all subsets $B \subseteq V^{-m}$ and for all values $x_c$ and $b$, we have:
\begin{equation}\label{eqn:context-independent}
 I(Y;X_m|B=b,X_c=x_c) = I(Y;X_m|B=b).
\end{equation}
\end{definition}
Context-dependent variables are thus the variables for which there exists a
conditioning set $B$ in which the information they bring about the output is
modified by the context variable. Context-independent variables are the
variables that, in all conditionings $B=b$, bring the same amount of
information about the output whether the value of the context is known or not.
This definition is meant to be as general as possible. Other more specific
definitions of context-dependence are as follows:\vspace{-0.5cm}

{\small
\begin{eqnarray}
&&\begin{split}
&\hspace{-3em}\exists B \subseteq V^{-m}, b, x^1_c, x^2_c:\\
&\hspace{-2em}I(Y;X_m|X_c=x^1_c,B=b) \neq I(Y;X_m|X_c=x^2_c,B=b),
 \end{split}
 \label{altcond1}\\
&&\begin{split}
&\hspace{-3em}\exists B \subseteq V^{-m},x_c:\\
&\hspace{-2em}I(Y;X_m|X_c=x_c,B) \neq I(Y;X_m|B),
\end{split}
\label{altcond2}\\
&&\begin{split}
&\hspace{-3em}\exists B\subseteq V^{-m},b:\\
&\hspace{-2em}I(Y;X_m|X_c,B=b) \neq I(Y;X_m|B=b),
\end{split}
\label{altcond3}\\
&&\begin{split}
&\hspace{-3em}\exists B\subseteq V^{-m}:\\
&\hspace{-2em}I(Y;X_m|X_c,B) \neq I(Y;X_m|B).
\end{split}
\label{altcond4}
\end{eqnarray}
}
These definitions all imply context-dependence as defined in Definition~\ref{defcontextdependent} but the converse is in general not true. For example, Definition~(\ref{altcond1}) misses problems where the context makes some otherwise irrelevant variable relevant but where the information brought by this variable about the output is exactly the same for all values of the context. A variable that satisfies Definition~(\ref{eqn:context-dependent}) but not Definition~(\ref{altcond2}) is given in example~\ref{example1}. This example can be easily adapted to show that both Definitions~(\ref{altcond3}) and (\ref{altcond4}) are more specific than Definition~(\ref{eqn:context-dependent}) (by swapping the roles of $X_c$ and $X_2$).

\begin{example}\label{example1}
  This artificial problem is defined by two input variables $X_1$ and $X_2$, an output $Y$, and a context $X_c$. $X_1$, $X_2$, and $X_c$ are binary variables taking their values in $\{0,1\}$, while $Y$ is a quaternary variable taking its values in $\{0,1,2,3\}$. All combinations of values for $X_1$, $X_2$, and $X_c$ have the same probability of occurrence $0.125$ and the conditional probability $P(Y|X_1,X_2,X_C)$ is defined by the two following rules:
  \begin{itemize}
    \item If $X_2=X_c$ then $Y=X_1$ with probability 1.
    \item If $X_2\neq X_c$ then $Y=2$ with probability $0.5$ and $Y=3$ with probability $0.5$.
  \end{itemize}
The corresponding data table is given in Appendix A. For
this problem, it is easy to show that $I(Y;X_1|X_2=0,X_c=0)=1$ and
that $I(Y;X_1|X_2=0)=0.5$, which means
condition~(\ref{eqn:context-dependent}) is satisfied and $X_1$ is thus
context-dependent to $Y$ with respect to $X_c$ according to our
definition. On the other hand, we can show that:
\begin{eqnarray*}
&& I(Y;X_1|X_c=x_c)=I(Y;X_1)=0.5\\
&& I(Y;X_1|X_2,X_c=x_c)=I(Y;X_1|X_2)=0.5,
\end{eqnarray*}
for any $x_c\in\{0,1\}$, which means that condition~(\ref{altcond2}) can not be satisfied for $X_1$.


\end{example}



To simplify the notations, the context variable was assumed to be a
separate variable not belonging to the set of inputs $V$. It can
however be considered as an input variable, whose own relevance to $Y$
(with respect to $V\cup \{X_c\}$) can be assessed as for any other
input.  Let us examine the impact of the nature of this variable on
context-dependence. First, it is interesting to note that the
definition of context-dependence is not symmetric. A variable $X_m$
being context-dependent to $Y$ with respect to $X_c$ does not imply
that the variable $X_c$ is context-dependent to $Y$ with respect to
$X_m$.\footnote{But this would be the case if we had adopted
definition (\ref{altcond4}).} Second, the context variable does not
need to be marginally relevant for some variable to be
context-dependent, but it needs however to be relevant to $Y$ with
respect to $V$. Indeed, we have the following theorem (proven in Appendix B):

\begin{theorem}\label{theo1} $X_c$ is irrelevant to $Y$ with respect to $V$ iff all variables in $V$ are context-independent to $Y$ with respect to $X_c$ (and $V$) and $I(Y;X_c)=0$.
\end{theorem}
As a consequence of this theorem, there is no interest in looking for
context-dependent variables when the context itself is not relevant.




\paragraph{Characterizing context-dependent variables.}

Contextual analyses need to focus only on context-dependent variables since,
by definition, context-independent variables are unaffected by the context:
their relevance status (relevant or irrelevant), as well as the information
they contain about the output, remain indeed unchanged whatever the
context.


Context-dependent variables may be affected in several directions by
the context, depending both on the conditioning subset $B$ and on the
value $x_c$ of the context. Given a context-dependent variable $X_m$,
a subset $B$ and some values $b$ and $x_c$ such that
$I(Y;X_m|B=b,X_c=x_c)\neq I(Y;X_m|B=b)$, the effect of the context can
either be an increase of the information brought by $X_m$
($I(Y;X_m|B=b,X_c=x_c) > I(Y;X_m|B=b)$) or a decrease of this
information ($I(Y;X_m|B=b,X_c=x_c) < I(Y;X_m|B=b)$). Furthermore, for
a given variable $X_m$, the direction of the change can differ from
one context value $x_c$ to another (at fixed $B$ and $b$) but also
from one conditioning $B=b$ to another (for a fixed context
$x_c$). Example~\ref{example2} below illustrates this latter
case. This observation makes a global characterization of the effect
of the context on a given context-dependent variable difficult. Let us
nevertheless mention two situations where such global characterization
is possible:

\begin{definition}
A context-dependent variable $X_m\in V$ is {\em context-complementary} (in a context $x_c$) iff for all
$B \subseteq V^{-m}$ and $b$, we have $I(Y;X_m|B=b,X_c=x_c) \ge
I(Y;X_m|B=b)$.
\end{definition}
\begin{definition}
A context-dependent variable $X_m\in V$ is {\em context-redundant} (in a context $x_c$)  iff for all
$B \subseteq V^{-m}$ and $b$, we have $I(Y;X_m|B=b,X_c=x_c) \le
I(Y;X_m|B=b)$.
\end{definition}

Context-complementary and redundant variables are variables that
always react in the same direction to the context and thus can be
characterized globally without loss of
information. Context-complementary variables are variables that bring
complementary information about the output with respect to the
context, while context-redundant variables are variables that are
redundant with the context. Note that context-dependent variables that
are also irrelevant to $Y$ are always context-complementary, since the
context can only increase the information they bring about the
output. Context-dependent variables that are relevant to $Y$ however
can be either context-complementary, context-redundant, or
uncharacterized. A context-redundant
variable can furthermore become irrelevant to $Y$
as soon as $I(Y;X_m|B=b,X_c=x_c)=0$ for all $B$, $b$, and $x_c$.

\begin{example}\label{example2}
As an illustration, in the problem of Example~\ref{example1}, $X_1$
and $X_2$ are both relevant and context-dependent variables. $X_1$ can
not be characterized globally since we have simultaneously:
\begin{eqnarray*}
  I(Y;X_1|X_2=0,X_c=x_c)&>&I(Y;X_1|X_2=0)\\
  I(Y;X_1|X_2=1,X_c=x_c)&<&I(Y;X_1|X_2=1),
\end{eqnarray*}
for both $x_c=0$ and $x_c=1$. $X_2$ is however context-complementary
as the knowledge of $X_c$ always increases the information it contains
about $Y$.
\end{example}

\paragraph{Related works.}

Several authors have studied interactions between variables in the context of
supervised learning. They have come up with various interaction definitions and
measures, e.g., based on multivariate mutual information
\citep{mcgill1954multivariate,jakulin2003analyzing}, conditional mutual
information \citep{jakulin2005machine, van2011two}, or variants thereof
\citep{brown2009new, brown2012conditional}. There are several differences
between these definitions and ours. In our case, the context variable has a
special status and as a consequence, our definition is inherently asymmetric,
while most existing variable interaction measures are symmetric. In addition,
we are interested in detecting any information difference occurring in a given
context (i.e., for a specific value of $X_c$) and for any conditioning subset
$B$, while most interaction analyses are interested in average and/or
unconditional effects. For example, \citep{jakulin2003analyzing} propose as a
measure of the interaction between two variables $X_1$ and $X_2$ with respect
to an output $Y$ the multivariate mutual information, which is defined as
$I(Y;X_1;X_2)=I(Y;X_1)-I(Y;X_1|X_2)$. Unlike our definition, this measure can
be shown to be symmetric with respect to its arguments. Adopting this measure
to define context-dependence would actually amount at using
condition~(\ref{altcond4}) instead of condition~(\ref{defcontextdependent}),
which would lead to a more specific definition as discussed earlier in this
section.

The closest work to ours in this literature is due to
\cite{turney1996identification}, who proposes a definition of
context-sensitivity that is very similar to our definition of
context-dependence. Using our notations, \cite{turney1996identification}
defines a variable $X_m$ as weakly context-sensitive to the variable $X_c$ if
there exist some subset $B\subseteq V^{-m}$ and some values $y$, $x_m$, $b$,
and $x_c$ such that these two conditions hold:
\begin{align*}
\resizebox{0.48\textwidth}{!}{$p(Y=y|X_m=x_m,X_c=x_c,B=b)\neq p(Y=y|X_m=x_m,B=b),$}\\
\resizebox{0.48\textwidth}{!}{$p(Y=y|X_m=x_m,X_c=x_c,B=b)\neq p(Y=y|X_c=x_c,B=b).$}
\end{align*}
$X_m$ is furthermore defined as strongly context-sensitive to $X_c$ if $X_m$ is
weakly sensitive to $X_c$, $X_m$ is marginally relevant,and $X_c$ is not
marginally relevant. These two definitions do not exactly coincide with ours
and they have two drawbacks in our opinion. First, they do not consider that a
perfect copy of the context is context-sensitive, which we think is
counter-intuitive. Second, while strong context-sensitivity is asymmetric, the
constraints about the marginal relevance of $X_m$ and $X_c$ seems also
unnatural.

Our work is also somehow related to several works in the graphical model
literature that are concerned with context-specific independences between
random variables \citep[see e.g.][]{boutilier1996,zhang99}. \cite{boutilier1996}
define two variables $Y$ and $X_m$ as contextually independent given some
$B\subseteq V^{-m}$ and a context value $x_c$ as soon as
$I(Y;X_m|B,X_c=x_c)=0$. When $B\cup\{X_m,X_c\}$ are the parents of node $Y$ in
a Bayesian network, then such context-specific independences can be exploited
to simplify the conditional probability tables of node $Y$ and to speed up
inferences. \cite{boutilier1996}'s context-specific independences will be captured by our
definition of context-dependence as soon as $I(Y;X_m|B)>0$. However, our
framework is more general as we want to detect any context dependencies, not
only those that lead to perfect independences in some context.

\section{\textsc{Context analysis with random forests}}
\label{sec:contextual-relevance-trees}

In this section, we show how to use variable importances derived from Random
Forests first to identify context-dependent variables
(Section~\ref{sec:identification}) and then to characterize the effect of the
context on the relevance of these variables
(Section~\ref{sec:charact}). Derivations in this section are based on the
theoretical characterization of variable importances provided in
\citep{louppe2013understanding}, which is briefly reminded in
Section~\ref{backgroundtrees}. Section~\ref{sec:in-practice} discusses
practical considerations and Section~\ref{sec:generalisation} shows how to
generalize our results to other impurity measures.

\subsection{Variable importances}\label{backgroundtrees}


Within the random forest framework,
\cite{breiman2001random} proposed to evaluate the importance of
a variable $X_m$  for predicting  $Y$ by adding up the weighted impurity decreases
for all nodes $t$ where $X_m$ is used, averaged over all $N_T$ trees
in the forest:
\begin{equation} \label{eqn:impfini}
  Imp(X_m) = \frac{1}{N_T} \sum_{T} \sum_{t\in T:v(s_t)=X_m} p(t) I(Y;X_m|t)
\end{equation}
where $v(s_t)$ is the variable used in the split $s_t$ at node $t$, $p(t)$ is the proportion of samples reaching $t$ and $I$ is the mutual information.

According to \cite{louppe2013understanding}, for any ensemble of fully
developed trees in asymptotic learning sample size conditions, the Mean Decrease Impurity (MDI)
importance~(\ref{eqn:impfini}) can be shown to be equivalent to
\begin{equation} \label{eqn:impasymp}
  Imp(X_m) = \sum_{k=0}^{p-1} \frac{1}{C^k_p}\frac{1}{p-k} \sum_{B\in{\cal P}_k(V^{-m})} I(Y;X_m|B),
\end{equation}
where  ${\cal P}_k(V^{-m})$ denotes
the set of subsets of $V^{-m}$ of size $k$.
Most notably, it can be shown \citep{louppe2013understanding}  that this
measure is zero for a variable $X_m$ iff $X_m$ is irrelevant to $Y$ with
respect to $V$. It is therefore well suited for identifying relevant features.

\subsection{Identifying context-dependent variables}\label{sec:identification}

Theorem~\ref{theo1} shows that if the context variable $X_c$ is irrelevant,
then it can not interact with the input variables and thus modify their
importances. This observation suggests to perform, as a preliminary test, a
standard random forest variable importance analysis using all input variables
and the context in order to check the relevance of the latter. If the context
variable does not reveal to be relevant, then, there is no hope to find
context-dependent variables.

Intuitively, identifying context-dependent variables seems similar to
identifying the variables whose importance is globally modified when the
context is known.
Therefore, one first straightforward approach to identify context-dependent
variables is to build a forest per value $X_c=x_c$ of the context variable, i.e.,
using only the data samples for which $X_c=x_c$ , and also
globally, i.e. using all samples and not including the context among the
inputs. Then it consists in deriving from these models an importance score for
each value of the context, as well as a global importance score.
Context-dependent variables are then the variables whose global importance score
differs from the contextual importance scores for at least one value of the
context.

More precisely, let us denote by $Imp(X_m)$ the global score of a variable
$X_m$ computed using (\ref{eqn:impfini}) from all samples and by
$Imp(X_m|X_c=x_c)$ its importance score as computed according to
(\ref{eqn:impfini}) using only those samples such that $X_c=x_c$. With this
approach, a variable would be declared as context-dependent as soon as there
exists a value $x_c$ such that $Imp(X_m)\neq Imp(X_m|X_c=x_c)$.

Although straightforward, this approach has several drawbacks. First, in the
asymptotic setting of Section~\ref{backgroundtrees}, it is not
guaranteed to find all context-dependent variables. Indeed, asymptotically, it
is easy to show from (\ref{eqn:impasymp}) that $Imp(X_m)-Imp(X_m|X_c=x_c)$ can
be written as:\vspace{-0.5cm}

{\small
\begin{eqnarray}
  Imp^{x_c}(X_m) &\triangleq& Imp(X_m)-Imp(X_m|X_c=x_c)\\
  &=& \begin{aligned} &\sum_{k=0}^{p-1} \frac{1}{C_k^p} \frac{1}{p-k} \sum_{B\in{\cal P}_k(V^{-m})} \\ &\hookrightarrow  (I(Y;X_m|B)-I(Y;X_m|B,X_c=x_c)). \end{aligned} \label{mdidiffcontext}
\end{eqnarray}}%
\vspace{-5mm}

Example~\ref{example1} shows that $I(Y;X_m|B)$ can be equal to
$I(Y;X_m|B,X_c=x_c)$ for a context-dependent variable. Therefore we have the
property that if there exists an $x_c$ such that $Imp^{x_c}(X_m)\neq 0$, then
the variable is context-dependent but the opposite is unfortunately not true.
Another drawback of this approach is that in the finite case, we do not have
the guarantee that the different forests will have explored the same
conditioning sets $B$ and therefore, even assuming that the learning sample is
infinite (and therefore that all mutual informations are perfectly estimated),
we lose the guarantee that $Imp^{x_c}(X_m)\neq 0$ for a given $x_c$ implies
context-dependence.

To overcome these issues, we propose the following new importance score to identify context-dependent variables:\vspace{-5mm}

{\small
\begin{equation}\label{impabs}
\begin{aligned}  Imp^{|x_c|}(X_m) \triangleq &  \frac{1}{N_T} \sum_T \sum_{t\in T:v(s_t)=X_m} p(t) \\
  &\hookrightarrow  |I(Y;X_m|t)-I(Y;X_m|t,X_c=x_c)| \end{aligned}
\end{equation}}
\vspace{-5mm}

This score is meant to be computed from a forest of totally randomized trees
built from all samples, not including the context variable among the inputs. At
each node $t$ where the variable $X_m$ is used to split, one needs to compute
the absolute value of the difference between the mutual information between $Y$
and $X_m$ estimated from all samples reaching that node and the mutual
information between $Y$ and $X_m$ estimated only from the samples for which
$X_c=x_c$. The same forest can then be used to compute $Imp^{|x_c|}(X_m)$ for
all $x_c$. A variable $X_m$ is then declared context-dependent as soon
as there exists an $x_c$ such that $Imp^{|x_c|}(X_m)>0$.


Let us show that this measure is sound. In asymptotic conditions,
i.e., with an infinite number of trees, one can show from
(\ref{impabs}) that $Imp^{|x_c|}(X_m)$ becomes:\vspace{-0.2cm}
{\small
\begin{equation*}
	\begin{split}
	Imp^{|x_c|}(X_m)  = &\sum_{k=0}^{p-1} \frac{1}{C^k_p} \frac{1}{p-k} \sum_{B\in{\cal P}_k(V^{-m})} \sum_{b\in {\cal B}} P(B=b)  \\
	&\hspace{-1.5em}\hookrightarrow  \left | I(Y;X_m|B=b)  - I(Y;X_m|B=b;X_c=x_c) \right | .
	\end{split}
\end{equation*}}
\vspace{-5mm}

Asymptotically, this measure has now the very desirable property to not miss any context-dependent variable as formalized in the next theorem (the proof is in Appendix C).
\begin{theorem}\label{theo2}
A variable $X_m \in V$ is context-independent to $Y$ with respect to $X_c$ iff $Imp^{|x_c|}(X_m)=0$ for all $x_c$.
\end{theorem}
Given that the absolute differences are computed at each tree node, this measure also continues to imply context-dependence in the case of finite forests and infinite learning sample size. The only difference with the infinite forests is that only some conditionings $B$ and values $b$ will be tested and therefore one might miss the conditionings that are needed to detect some context-dependent variables.

\subsection{Characterizing context-dependent variables}\label{sec:charact}

Besides identifying context-dependent variables, one would want to characterize
their dependence with the context as precisely as possible. As discussed earlier,
irrelevant variables (i.e, such that
$Imp(X_m)=0$) that are detected as context-dependent do not need much effort to
be characterized since the context can only increase their importance. All
these variables are therefore context-complementary.

Identifying the context-complementary and context-redundant variables among the
relevant variables that are also context-dependent can in principle be done by
simply comparing the absolute value of $Imp^{x_c}(X_m)$ with
$Imp^{|x_c|}(X_m)$, as formalized in the following theorem (proven in
Appendix D).\\

\vspace{-\topsep}
\begin{theorem}\label{theo3} If $|Imp^{x_c}(X_m)| = Imp^{|x_c|}(X_m)$ for a context-dependent variable $X_m$, then $X_m$ is context-complementary if $Imp^{x_c}(X_m) < 0$ and context-redundant if $Imp^{x_c}(X_m) > 0$.
\end{theorem}
\vspace{-\topsep}
This result allows to identify easily the context-complementary and
context-redundant variables. In addition, if, for a context-redundant variable
$X_m$, we have $Imp^{|x_c|}(X_m)=Imp^{x_c}(X_m)=Imp(X_m)$, then this variable is irrelevant in
the context $x_c$.

Then it remains to characterize the context-dependent variables that are neither
context-complementary nor context-redundant. It would be interesting to be able
to also characterize them according to some sort of average effect of the
context on these variables. Similarly as the common use of importance $Imp(X_m)$ to
rank variables from the most to the less important, we propose to use the
importance $Imp^{x_c}(X_m)$ to characterize the average global effect of
context $x_c$ on the variable $X_m$. Given the asymptotic formulation of this
importance in Equation~(\ref{mdidiffcontext}), a negative value of
$Imp^{x_c}(X_m)$ means that $X_m$ is essentially complementary with the
context: in average over all conditionings, it brings more information about
$Y$ in context $x_c$ than when ignoring the context. Conversely, a positive
value of $Imp^{x_c}(X_m)$ means that the variable is essentially redundant with
the context: in average over all conditionings, it brings less information
about $Y$ than when ignoring the context. Ranking the context-dependent
variables according to $Imp^{x_c}(X_m)$ would then give at the top the
variables that are the most complementary with the context and at the bottom
the variables that are the most redundant.

Note that, like $Imp^{|x_c|}(X_m)$, it is preferable to estimate
$Imp^{x_c}(X_m)$ by using the following formula rather than to estimate it from
two forests by subtracting $Imp(X_m)$ and $Imp(X_m|X_c=x_c)$:
\begin{equation}\label{eqn:impxcnodebynode}
  \begin{split} Imp_s^{x_c}(X_m)=  & \frac{1}{N_T} \sum_T \sum_{t\in T:v(s_t)=X_m} p(t) \\ & \hookrightarrow  (I(Y;X_m|t)-I(Y;X_m|t,X_c=x_c)) \end{split}
\end{equation}
This estimation method has the same asymptotic form as
$Imp(X_m)-Imp(X_m|X_c=x_c)$ given in Equation~(\ref{mdidiffcontext}) but, in
the finite case, it ensures that the same conditionings are used for both
mutual information measures. Note that in some applications, it is interesting
also to have a global measure of the effect of the context. A natural adaptation
of (\ref{eqn:impxcnodebynode}) to obtain such global measure is as follows:
\begin{equation*}
\begin{split} Imp^{X_c}(X_m) \triangleq & \frac{1}{N_T} \sum_T \sum_{t\in T:v(s_t)=X_m} p(t) \\ & \hookrightarrow (I(Y;X_m|t)-I(Y;X_m|t,X_c))\end{split}
\end{equation*}
\vspace{-5mm}

which, in asymptotic sample and ensemble of trees size conditions, gives the following formula:\vspace{-3mm}
\begin{equation*}
 \begin{split}Imp^{X_c}(X_m) =  & \sum_{k=0}^{p-1} \frac{1}{C_k^p} \frac{1}{p-k} \sum_{B\in{\cal P}_k(V^{-m})} \\ & \hookrightarrow  (I(Y;X_m|B)-I(Y;X_m|B,X_c)).\end{split}
\end{equation*}
\vspace{-5mm}

If $Imp^{X_c}(X_m)$ is negative then the context variable $X_c$ makes variable
$X_m$ globally more informative ($X_c$ and $X_m$ are complementary with
respect to $Y$ and $V$). If $Imp^{X_c}(X_m)$ is positive, then the context
variable $X_c$ makes variable $X_m$ globally less informative ($X_c$ and $X_m$
are redundant with respect to $Y$ and $V$).

\subsection{In practice}
\label{sec:in-practice}



As a recipe when starting a context analysis, we suggest first to build a single forest using all input variables $X_m$ (but not the context $X_c$) and then to compute from this forest all importances defined in the previous section: the global importances $Imp(X_m)$ and the different contextual importances, $Imp_s^{x_c}(X_m)$, $Imp^{|x_c|}(X_m)$, and $Imp^{X_c}(X_m)$, for all variables $X_m$ and context values $x_c$.

Second, variables satisfying the context-dependence criterion, i.e., such that $Imp^{|x_c|}(X_m)>0$ for at least one $x_c$, can be identified from the other variables. Among context-dependent variables, an equality between $|Imp_s^{x_c}(X_m)|$ and $Imp^{|x_c|}(X_m)$ highlights that the context-dependent variable $X_m$ is either context-complementary or context-redundant (in $x_c$) depending on the sign of $Imp_s^{x_c}(X_m)$. Finally, the remaining context-dependent variables can be ranked according to $Imp_s^{x_c}(X_m)$ (or $Imp^{X_c}(X_m)$ for a more global analysis).

Note that, because mutual informations will be estimated from finite training sets, they will be generally non zero even for independent variables, leading to false positives in the identification of context-dependent variables. In practice, one could instead identify context-dependent variables by using a test $Imp^{|x_c|}(X_m)> \epsilon$ where $\epsilon$ is some cut-off value greater than 0. In practice, the determination of this cut-off can be very difficult. In our experiments, we propose to turn the importances $Imp^{|x_c|}(X_m)$ into $p$-values by using random permutations. More precisely, 1000 scores $Imp^{|x_c|}(X_m)$ will be estimated by randomly permuting the values of the context variable in the original data (so as to simulate the null hypothesis corresponding to a context variable fully independent of all other variables). A $p$-value will then be estimated by the proportion of these permutations leading to a score $Imp^{|x_c|}(X_m)$ greater than the score obtained on the original dataset.

   \begin{table}[htbp]
  {\scriptsize
    \centering
\begin{tabular}{c|ccc|c} \hline
$X_c$ & $X_1$ & $X_2$ & $X_3$ & $Y$\\ \hline
    0   &   0\g &   0   &   0   & \g2\\
    0   &   0\g &   0   &   1   & \g2\\
    0   &   0\g &   1   &   0   & \g2\\
    0   &   0\g &   1   &   1   & \g2\\
    0   &   1   & \g0   &   0   & \g0\\
    0   &   1   & \g0   &   1   & \g0\\
    0   &   1   & \g1   &   0   & \g1\\
    0   &   1   & \g1   &   1   & \g1\\ \hline
    1   &   0\g &   0   &   0   & \g2\\
    1   &   0\g &   0   &   1   & \g2\\
    1   &   0\g &   1   &   0   & \g2\\
    1   &   0\g &   1   &   1   & \g2\\
    1   &   1   &   0   & \g0   & \g0\\
    1   &   1   &   0   & \g1   & \g1\\
    1   &   1   &   1   & \g0   & \g0\\
    1   &   1   &   1   & \g1   & \g1\\ \hline
\end{tabular}}
\captionof{table}{Problem 1: Values of $X_c$, $X_1$, $X_2$, $X_3$, $Y$.}
\label{table:data1}
  \end{table}

   \begin{table}[htbp]
{\scriptsize
    \centering
\begin{tabular}{l|lll} \hline
        & $X_1$ & $X_2$ & $X_3$ \\ \hline
$Imp(X_m)$        & 1.0 & 0.125 & 0.125\\ 
  $Imp(X_m|X_c=0)$    & 1.0 & 0.5 & 0.0\\ 
  $Imp(X_m|X_c=1)$    & 1.0 & 0.0 & 0.5\\ 
$Imp^{|0|}(X_m)$  & 0.0 & 0.375 & 0.125 \\ 
  $Imp^{0}(X_m)$    & 0.0 & -0.375& 0.125 \\ 
  $Imp^{|1|}(X_m)$  & 0.0 & 0.125 & 0.375 \\ 
  $Imp^{1}(X_m)$  & 0.0 & 0.125 & -0.375 \\ 
  $Imp^{X_c}(X_m)$  & 0.0 & -0.125  & -0.125  \\ \hline 
\end{tabular}}
\captionof{table}{Problem 1: Variable importances as computed analytically using asymptotic formulas. Note that $X_1$ is context-independent
  and $X_2$ and $X_3$ are context-dependent.}\label{table:impexp1} 
  \vspace{-5mm}
  \end{table}

\begin{table*}[t]
{\scriptsize
\begin{tabular}{l|llllllll} \hline
						& $X_1$ & $X_2$ & $X_3$	& $X_4$ & $X_5$ & $X_6$ & $X_7$ & $X_8$ \\ \hline
$Imp(X_m)$				&0.5727		&0.7514		&0.5528		&0.687 		&0.1746		&0.0753		&0.1073		&0.0  \\
$Imp(X_m|X_c=0)$		&0.4127 	&0.5815		&0.5312		&0.5421		&0.6566		&0.2258		&0.372		&0.0\\
$Imp(X_m|X_c=1)$		&0.6243		&0.8057		&0.5577		&0.7343		&0.0		&0.0		&0.0		&0.0\\ \hline 
$Imp^{|0|}(X_m)$ 		&0.2263		&0.2431		&0.1181		&0.2241		&0.4139		&0.1961		&0.2861		&0.0\\
$Imp^{|1|}(X_m)$ 		&0.0987		&0.0611		&0.021		&0.0736		&0.1746		&0.0753		&0.1073		&0.0\\ \hline
$Imp^{0}(X_m)$			&0.2179		&0.2422		&0.1111		&0.2190		&-0.3839 	&-0.1389	&-0.2346	&0.0\\ 
$Imp^{1}(X_m)$			&-0.0516	&-0.0543	&-0.0049	&-0.0473	&0.1746		&0.0753		&0.1073		&0.0\\ \hline 
\end{tabular}}

\caption{Problem 2: Variable importances as computed analytically using the asymptotic formulas for the different importance measures.\vspace{-2mm} 
}\label{table:impexp2}
\vspace{-0.2cm}
\end{table*}

\begin{table*}[t]
{\scriptsize
\begin{tabular}{c|c|c|cc|cc:cc|cc:cc} \hline
& &             $Imp(X_m)$  & \multicolumn{2}{c|}{$Imp(X_m|X_c=x_c)$}   &  \multicolumn{4}{c|}{$Imp^{|x_c|}(X_m)$} & \multicolumn{4}{c}{$Imp_s^{x_c}(X_m)$}\\
m &             & -     & $x_c=0$   & $x_c=1$             &  $x_c=0$ & pval   & $x_c=1$ & pval  & $x_c=0$ & pval & $x_c=1$ & pval \\ \hline
0 & age              & 0.2974 & 0.2942 & 0.2900 & 0.1505 &   0.899 & 0.1717 &   0.417 &  0.0032  &   0.938 &  0.0074 &   0.846 \\
1 & histologic-type  & 0.3513 & 0.1354 & 0.4005 & 0.2265 &\g 0.000 & 0.1183 &   0.121 &  0.2159  &\g 0.000 & -0.0492 &   0.331 \\
2 & degree-of-diffe  & 0.4415 & 0.3725 & 0.4070 & 0.1827 &   0.680 & 0.1724 &   0.689 &  0.0690  &   0.102 &  0.0345 &   0.398 \\
3 & bone             & 0.2452 & 0.2342 & 0.2220 & 0.1088 &   0.396 & 0.0845 &   0.904 &  0.0110  &   0.717 &  0.0232 &   0.410 \\
4 & bone-marrow      & 0.0188 & 0.0190 & 0.0131 & 0.0128 &   0.892 & 0.0105 &   0.980 & -0.0001  &   0.994 &  0.0057 &   0.682 \\
5 & lung             & 0.1677 & 0.1837 & 0.1420 & 0.1134 &   0.448 & 0.1079 &   0.397 & -0.0160  &   0.605 &  0.0257 &   0.373 \\
6 & pleura           & 0.1474 & 0.1132 & 0.1127 & 0.0613 &   1.000 & 0.1026 &   0.097 &  0.0342  &   0.179 &  0.0348 &   0.165 \\
7 & peritoneum       & 0.3171 & 0.2954 & 0.2084 & 0.0939 &   0.968 & 0.1516 &\g 0.000 &  0.0216  &   0.710 &  0.1087 &\g 0.000\\
8 & liver            & 0.2300 & 0.1844 & 0.2784 & 0.0888 &   0.966 & 0.1382 &   0.053 &  0.0456  &   0.134 & -0.0483 &   0.100 \\
9 & brain            & 0.0466 & 0.0334 & 0.0566 & 0.0403 &   0.173 & 0.0279 &   0.814 &  0.0131  &   0.693 & -0.0101 &   0.751 \\
10 & skin            & 0.0679 & 0.0310 & 0.0786 & 0.0426 &   0.922 & 0.0420 &   0.841 &  0.0369  &   0.107 & -0.0107 &   0.663 \\
11 & neck            & 0.2183 & 0.0774 & 0.2255 & 0.1562 &\g 0.000 & 0.0710 &   0.575 &  0.1409  &\g 0.000 & -0.0071 &   0.764 \\
12 & supraclavicular & 0.1701 & 0.1807 & 0.1344 & 0.0942 &   0.379 & 0.0738 &   0.884 & -0.0106  &   0.695 &  0.0357 &   0.136 \\
13 & axillar         & 0.1339 & 0.1236 & 0.0846 & 0.0748 &   0.214 & 0.0663 &   0.388 &  0.0103  &   0.795 &  0.0493 &   0.194 \\
14 & mediastinum     & 0.1826 & 0.1752 & 0.1613 & 0.1129 &   0.266 & 0.0867 &   0.853 &  0.0074  &   0.767 &  0.0213 &   0.404 \\
15 & abdominal       & 0.2558 & 0.2883 & 0.1512 & 0.1419 &   0.139 & 0.1526 &\g 0.028 & -0.0325  &   0.368 &  0.1046 &\g 0.003\\ \hline
\end{tabular}}
\caption{Problem 3: Importances as computed with a forest of 1000 totally randomized trees. The context is defined by the binary context feature \textit{Sex} ($Sex=0$ denotes \textit{female} and $Sex=1$ denotes \textit{male}). \textit{P-values} were estimated using 1000 permutations of the context variable. Grey cells highlight \textit{p-values} under the 0.05 threshold.
}
\label{table:forest-topresult}
\end{table*}

\subsection{Generalization to other impurity measures}\label{sec:generalisation}

All our developments so far have assumed a categorical output $Y$ and
the use of Shannon's entropy as the impurity measure. Our framework
however can be carried over to other impurity measures and thus in
particular also to a numerical output $Y$. Let us define a generic
impurity measure $i(Y|t)\geq 0$ that assesses the impurity of the
output $Y$ at a tree node $t$. The corresponding impurity decrease at
a tree node is defined as:
\begin{equation}\label{eqn:generic_reduction}
G(Y;X_m|t) = i(Y|t) - \sum_{{x_m}\in{\cal X}_m} p(t_{x_m})i(Y|t_{x_m})
\end{equation}
with $t_{x_m}$ denoting the successor node of $t$ corresponding to value
$x_m$ of $X_m$. By analogy with conditional entropy and mutual
information, let us define the population based measures $i(Y|B)$ and
$G(Y;X_m|B)$ for any subset of variables $B\subseteq V$ as follows:
\begin{eqnarray*}
  i(Y|B) &=&\sum_b P(B=b) i(Y|B=b)\\
  G(Y;X_m|B) &= &i(Y|B)-i(Y|B,X_m),
\end{eqnarray*}
where the first sum is over all possible combinations $b$ of values
for variables in $B$. Now, substituting mutual information $I$ for the
corresponding impurity decrease measure $G$, all our results above
remain valid, including Theorems 1, 2, and 3 (proofs are omitted for
the sake of space). It is important however to note that this
substitution changes the notions of both variable relevance and
context-dependence. Definition \ref{defcontextdependent} indeed
becomes:
\begin{definition}\label{defcontextdependentgeneral}
 A variable $X_m\in V$ is {\em context-dependent to $Y$ with respect
   to $X_c$} iff there exists a subset $B \subseteq V^{-m}$ and some
 values $x_c$ and $b$ such that $$G(Y;X_m|B=b,X_c=x_c) \ne
 G(Y;X_m|B=b).$$
\end{definition}
When $Y$ is numerical, a common impurity measure is variance, which
defines $i(Y|t)$ as the empirical variance $\mbox{var}[Y|t]$ computed
at node $t$. The corresponding $G(X_m;Y|B=b)$ and $G(X_m;Y|B=b,X_c=x_c)$
in Definition~\ref{defcontextdependentgeneral} are thus defined respectively as
\begin{eqnarray*}
  &\mbox{var}[Y|B=b]-\mathbb{E}_{X_m|B=b}[\mbox{var}[Y|X_m,B=b]]\, \text{and}
\end{eqnarray*}\vspace{-1mm}
\begin{equation*}
  \begin{split} & \mbox{var}[Y|B=b,X_c=x_c]\\
    & \hookrightarrow -\mathbb{E}_{X_m|B=b,X_c=x_c}[\mbox{var}[Y|X_m,B=b,X_c=x_c]] .\end{split}
\end{equation*}
We will illustrate the use of our framework in a regression
 setting with this measure in the next section.

\section{\textsc{Experiments}}
\label{sec:experiments}


\paragraph{Problem 1.}

The purpose of this first problem is to illustrate the different
measures introduced earlier. This artificial problem is defined by
three binary input variables $X_1$, $X_2$, and $X_3$, a ternary output
$Y$, and a binary context $X_c$. All samples are enumerated in
Table~\ref{table:data1} and are supposed to be equiprobable. By
construction, the output $Y$ is defined as $Y=2$ if $X_1=0$, $Y=X_2$
if $X_c=0$ and $X_1=1$, and $Y=X_3$ if $X_c=1$ and $X_1=1$.

Table~\ref{table:impexp1} reports all importance scores for
the three inputs. These scores were computed analytically using the
asymptotic formulas, not from actual experiments. Considering the
global importances $Imp(X_m)$, it turns out that all variables are
relevant, with $X_1$ clearly the most important variable and $X_2$ and
$X_3$ of smaller and equal importances. According to $Imp^{|0|}(X_m)$
and $Imp^{|1|}(X_m)$, $X_1$ is a context-independent variable, while
$X_2$ and $X_3$ are two context-dependent variables. This result is as
expected given the way the output is defined. For $X_2$ and $X_3$, we
have furthermore $Imp^{|x_c|}(X_m)=|Imp^{|x_c|}(X_m)|$ for both values
of $x_c$. $X_2$ is therefore context-complementary when $X_c=0$ and
context-redundant when $X_c=1$. Conversely, $X_3$ is context-redundant
when $X_c=0$ and context-complementary when $X_c=1$. $X_2$ is
furthermore irrelevant when $X_c=1$ (since
$Imp^{1}(X_2)=Imp^{|1|}(X_2)=Imp(X_2)$) and $X_3$ is irrelevant when
$X_c=0$ (since $Imp^{0}(X_3)=Imp^{|0|}(X_3)=Imp(X_3)$). The values of
$Imp^{X_c}(X_2)$ and $Imp^{X_c}(X_3)$ suggest that these two variables
are in average complementary.

\paragraph{Problem 2.}
\label{sec:s4.2}

This second experiment is based on an adaptation of the digit recognition
problem initially proposed in \cite{breiman1984classification} and reused in
\cite{louppe2013understanding}. The original problem contains 7 binary
variables ($X_1$,\ldots,$X_7$) and the output $Y$ takes its values in $\{0,1,\ldots,9\}$. Each input
represents the on-off status of one lightning segment of a
seven-segment indicator and is determined univocally from $Y$. To create an
artificial (binary) context, we created two copies of this dataset, the first
one corresponding to $X_c=0$ and the second one to $X_c=1$. The first dataset
was unchanged, while in the second one variables $X_5$, $X_6$, and $X_7$ were
turned into irrelevant variables. In addition, we included a new variable
$X_8$, irrelevant by construction in both contexts. The final dataset contains
320 samples, 160 in each context.






Table~\ref{table:impexp2} reports possible importance scores for all the
inputs. Again, these scores were computed analytically using the asymptotic
formulas. As expected, variable $X_8$ has zero importance in all cases. Also as
expected, variables $X_5$, $X_6$, and $X_7$ are all context-dependent
($Imp^{|x_c|}(X_m)>0$ for all of them). They are context-redundant (and even
irrelevant) when $X_c=1$ and complementary when
$X_c=0$. More surprisingly, variables $X_1$, $X_2$, $X_3$, and $X_4$ are also
context-dependent, even if their distribution is independent from the
context. This is due to the fact that these variables are complementary with
variables $X_5$, $X_6$, and $X_7$ for predicting the output. Their
context-dependence is thus a consequence of the context-dependence of $X_5$,
$X_6$, $X_7$. $X_1$, $X_2$, $X_3$, and $X_4$ are all almost redundant when
$X_c=0$ and complementary when $X_c=1$, which expresses the fact that they
provide more information about the output when $X_5$, $X_6$ and $X_7$ are
irrelevant ($X_c=1$) and less when $X_5$, $X_6$, and $X_7$ are relevant
($X_c=0$). Nevertheless, $X_8$ remains irrelevant in every situation.



\paragraph{Problem 3.}

We now consider bio-medical data from the \textit{Primary tumor} dataset. The objective of the
corresponding supervised learning problem is to predict the location
of a primary tumor in patients with metastases. It was downloaded from
the UCI repository \citep{Lichman2013UCI} and was collected by the
University Medical Center in Ljubljana, Slovenia. We restrict our
analysis to 132 samples without missing values. Patients are
described by 17 discrete clinical variables (listed in the first
column of Table~\ref{table:forest-topresult}) and the output is chosen among 22
possible locations. For this analysis, we use the patient gender as the
context variable.

Table~\ref{table:forest-topresult} reports variable importances computed
with 1000 totally randomized trees and their corresponding p-values.
According to the p-values of $Imp^{|x_c|}(X_m)$, two variables are
clearly emphasized for each context: importances of \textit{histologic-type} and
\textit{neck} both significantly decrease in the first context ($female$) and
importances of \textit{peritoneum} and \textit{abdominal} both significantly
decrease in the second context ($male$). While the biological relevance of
these finding needs to be verified, such dependences could not have
been highlighted from standard random forests importances.

Note that the same importances computed using the asymptotic formulas
are provided in Appendix E.  Importance
values are very similar, highlighting that finite forests provide good
enough estimates for this problem.

\paragraph{Problem 4.}

\begin{figure*}[t!]
\centering
\caption{Results for Problem 4. Each matrix represents significant
  context-dependent gene-gene interactions as found using $Imp^{|x_c|}$ in
  (a)(b) and $Imp^{x_c}$ in (c)(d), in GBM sub-type Mesenschymal in (a)(c) and
  Proneural in (b)(d). In (a) and (b), cells are colored according to
  $Imp_s^{x_c}$. In (c) and (d), cells are colored according to
  $Imp^{x_c}$. Positive (resp negative) values are in blue (resp. red) and
  highlight context-redundant (resp. context-complementary)
  interactions. Higher absolute values are darker.}
\label{fig:imp4}
\subfloat[$Imp^{|x_c=Mesenchymal|}$]{\label{imp4-a}\includegraphics[width=0.24\linewidth]{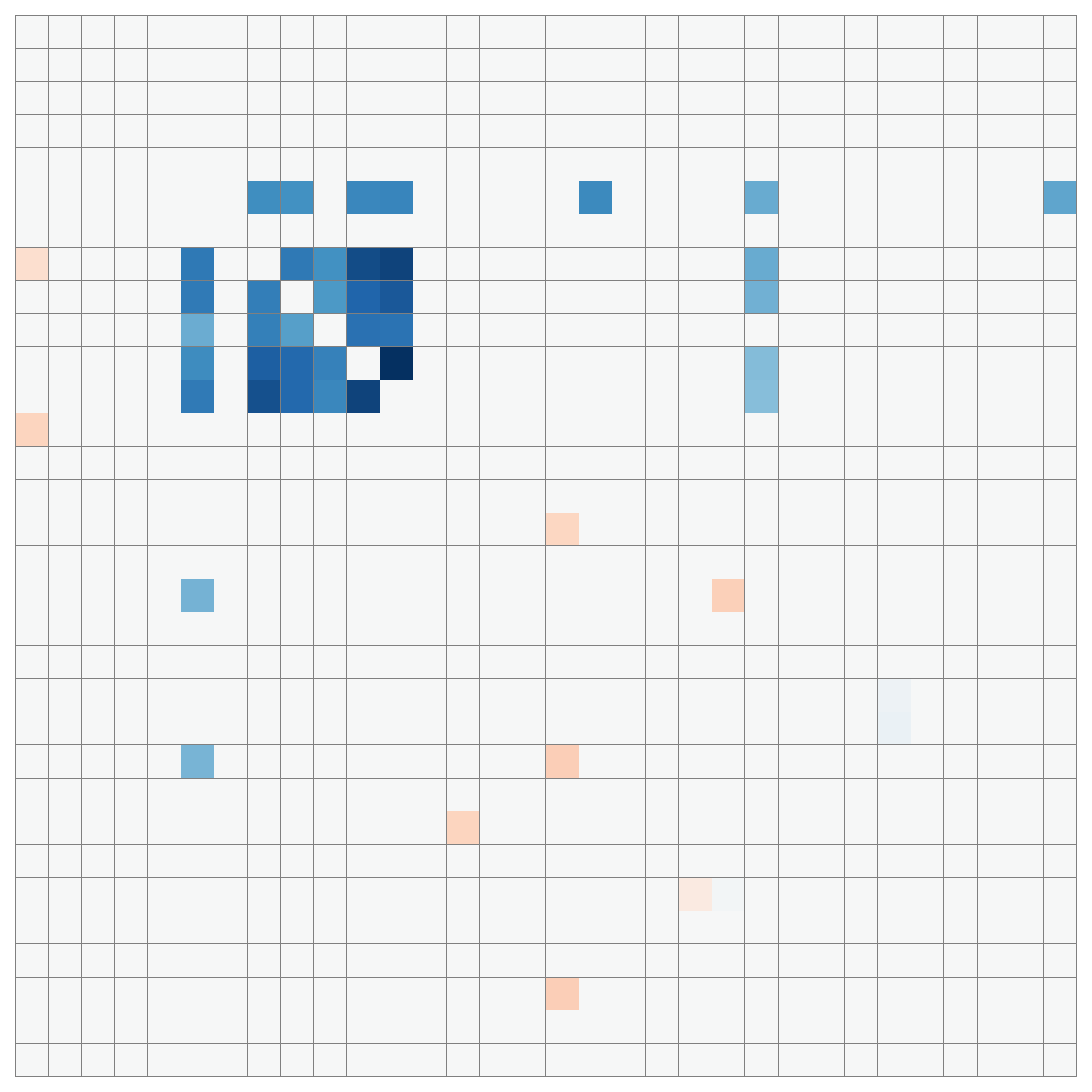}}
\subfloat[$Imp^{|x_c=Proneural|}$]{\label{imp4-b}\includegraphics[width=0.24\linewidth]{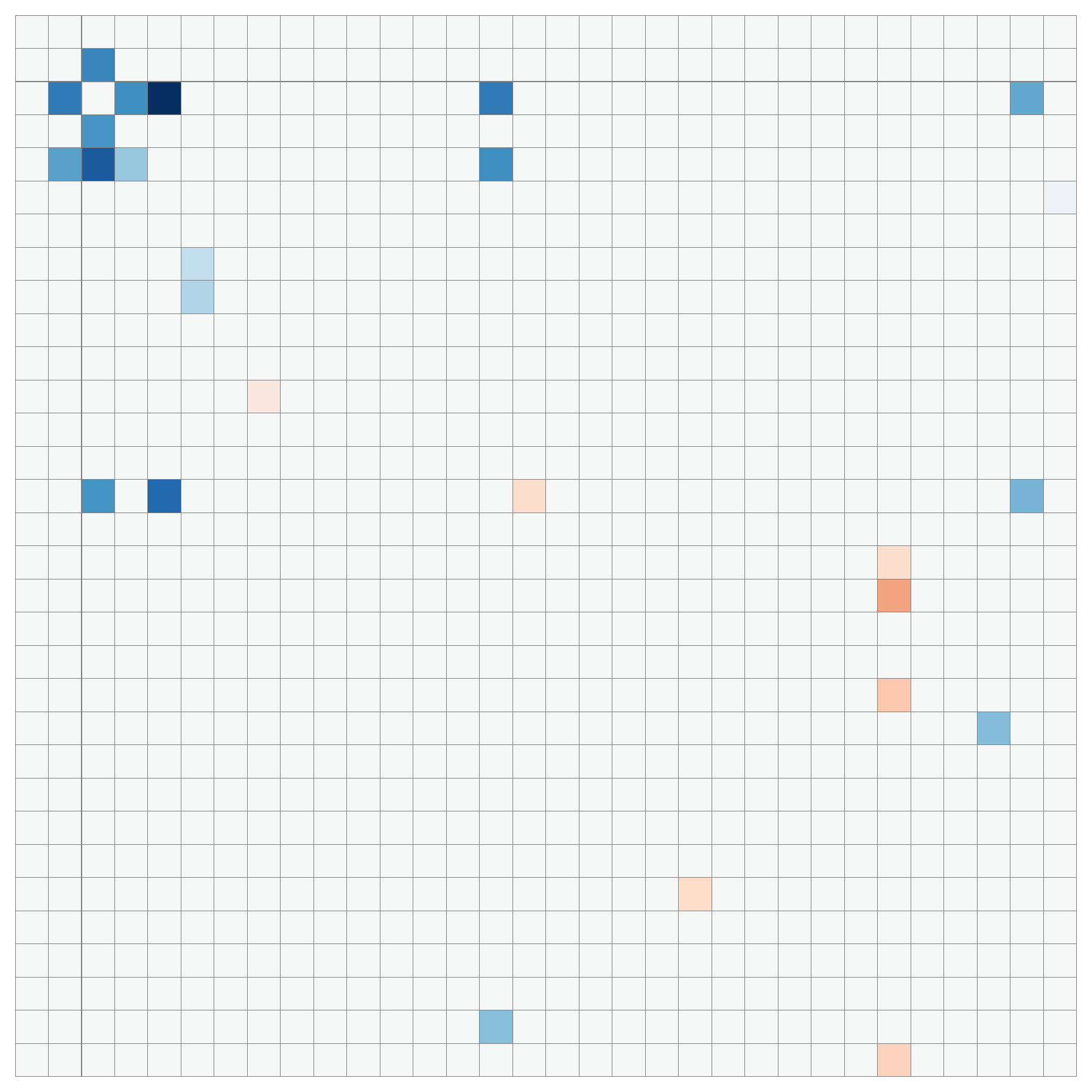}}
\subfloat[$Imp^{x_c=Mesenchymal}$]{\label{imp4-c}\includegraphics[width=0.24\linewidth]{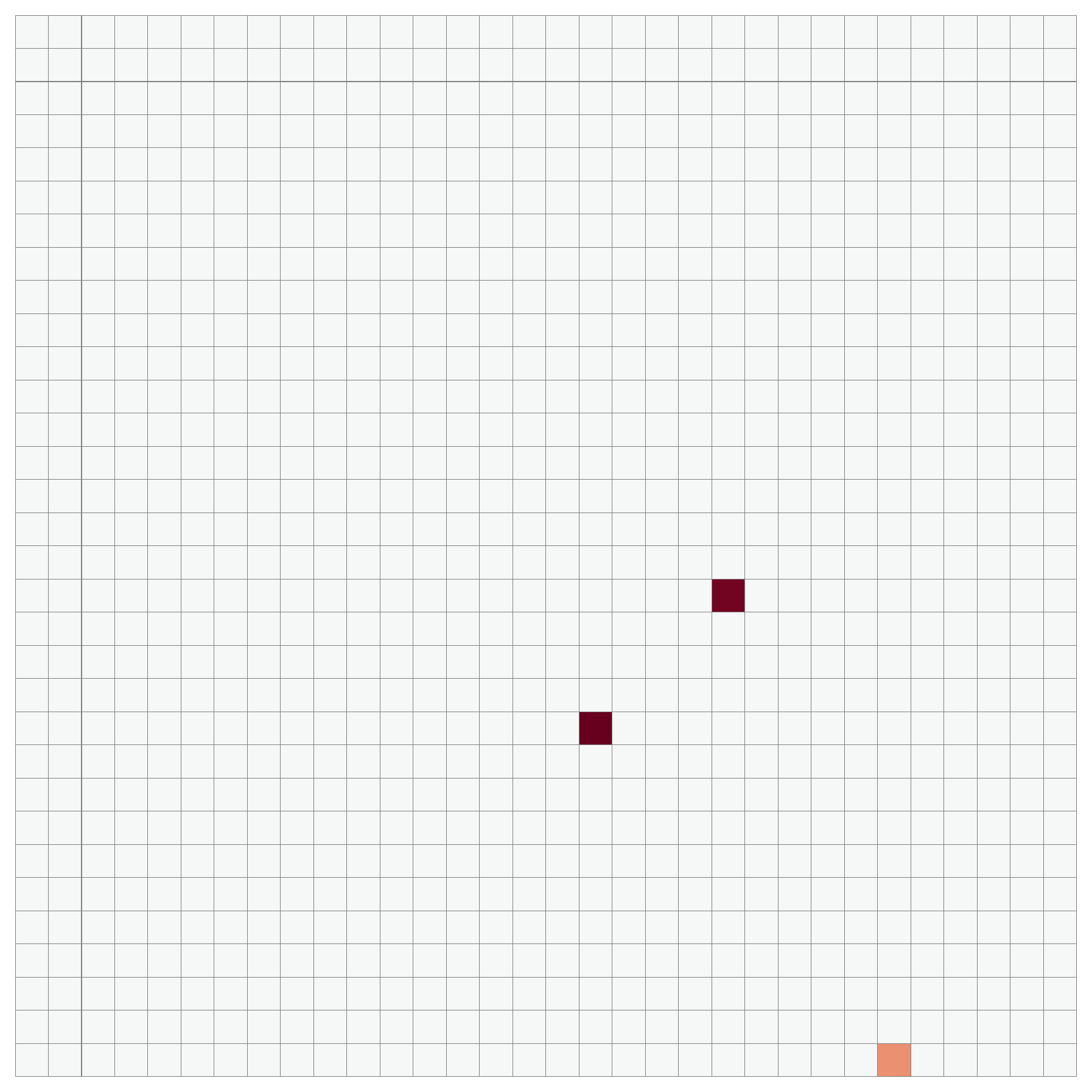}}
\subfloat[$Imp^{x_c=Proneural}$]{\label{imp4-d}\includegraphics[width=0.24\linewidth]{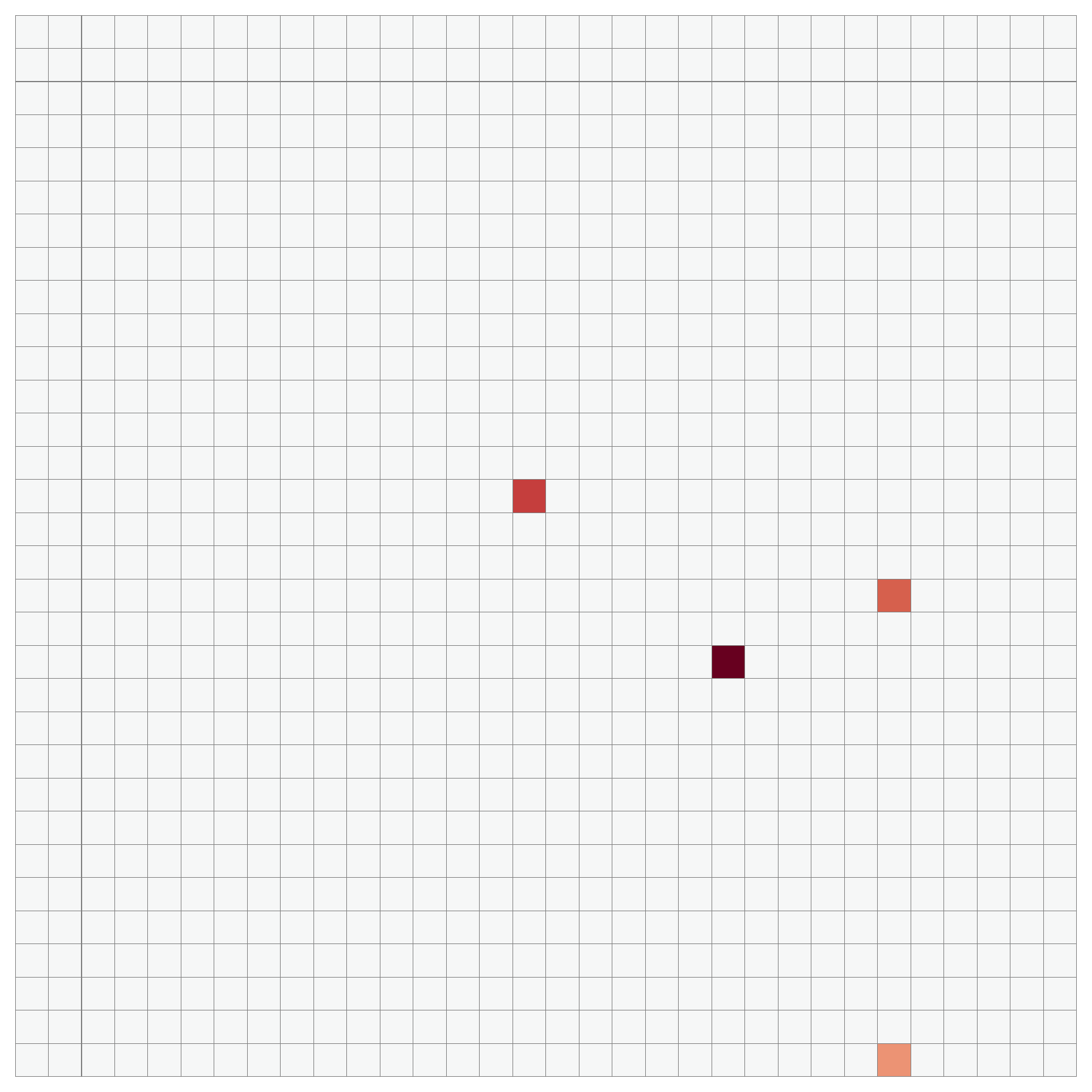}}
\includegraphics[width=0.045\linewidth]{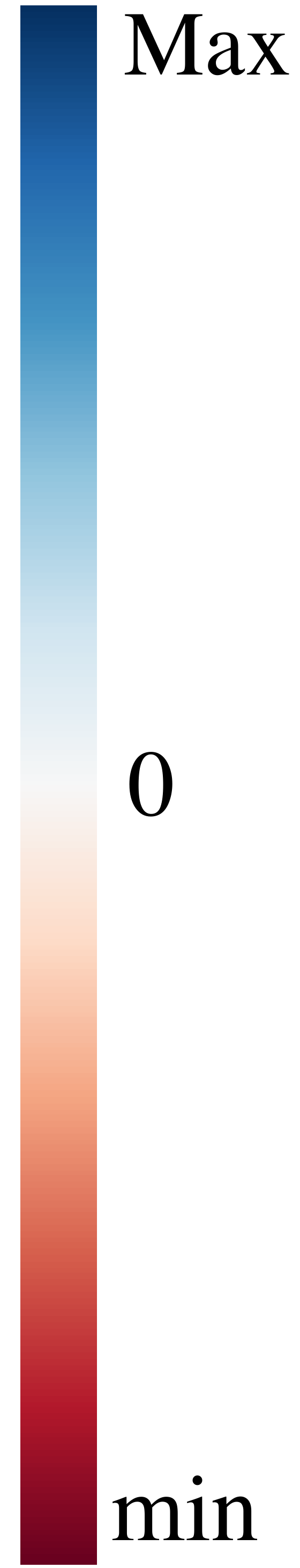}
\end{figure*}

As a last experiment, we consider a publicly available brain cancer
gene expression dataset \citep{verhaak2010integrated}. This dataset
collects measurements of mRNA expression levels of 11861 genes in 220
tissue samples from patients suffering from glioblastoma multiforme
(GBM), the most common form of malignant brain cancer in
adults. Samples are classified into four GBM sub-types: Classical,
Mesenchymal, Neural and Proneural. The interest of this dataset is to
identify the genes that play a central role in the development and
progression of the cancer and thus improve our understanding of this
disease. In our experiment, our aim is to exploit importance scores
to identify interactions between genes that are significantly affected
by the cancer sub-type considered as our context variable. This
dataset was previously exploited by \citet{mohan2014node}, who used it
to test a method based on Gaussian graphical models for detecting
genes whose global interaction patterns with all the other genes vary
significantly between the subtypes. This latter method can be
considered as gene-based, while our approach is link-based.

Following \citep{mohan2014node}, we normalized the raw data
using Multi-array Average (RMA) normalization. Then, the data was corrected for
batch effects using the software \textrm{ComBat} \citep{johnson2007adjusting}
and then $log2$ transformed. Following \citep{mohan2014node}, we focused our
analysis on only two GBM sub-types, Proneural (57 tissue samples) and
Mesenchymal (56 tissue samples), and on a particular set of 32 genes, which are
all genes involved in the TCR signaling pathway as defined in the Reactome
database \citep{matthews2009reactome}. The final dataset used in the
experiments below thus contains 113 samples, 57 and 56
for both context values respectively, and 32 variables.

To identify gene-gene interactions affected by the context, we performed a
contextual analysis as described in Section
\ref{sec:contextual-relevance-trees} for each gene in turn, considering each
time a particular gene as the target variable $Y$ and all other genes as the
set of input variables $V$. This procedure is similar to the procedure adopted
in the Random forests-based gene network inference method called GENIE3
\citep{huynh2010inferring}, that was the best performer in the DREAM5 network
inference challenge \citep{marbach2012wisdom}. Since gene expressions are
numerical targets, we used variance as the impurity measure (see Section
\ref{sec:generalisation}) and we built ensembles of 1000
totally randomized trees in all experiments.

The matrices in Figure~\ref{fig:imp4} highlight context-dependent interactions
found using different importance measures (detailed below). A cell $(i,j)$ of
these matrices corresponds to the importance of gene $j$ when gene $i$ is the
output (the diagonal is irrelevant). White cells correspond to non significant
context-dependencies as determined by random permutations of the context
variable, using a significance level of 0.05. Significant context-dependent
interactions in Figures~\ref{fig:imp4}(a) and (b) were determined using the
importance $Imp^{|x_c|}$ defined in (\ref{impabs}), which is the measure we
advocate in this paper. As a baseline for comparison, Figures~\ref{fig:imp4}(c)
and (d) show significant interactions as found using the more straightforward
score $Imp^{x_c}$ defined in (\ref{mdidiffcontext}). In
Figures~\ref{fig:imp4}(a) and (b) (resp. (c) and (d)), significant cells are
colored according to the value of $Imp_s^{x_c}$ defined in
(\ref{eqn:impxcnodebynode}). In Figures~\ref{fig:imp4}(c) and (d), they are
colored according to the value of $Imp^{x_c}$ in (\ref{mdidiffcontext})
instead. Blue (resp. red) cells correspond to positive (resp. negative) values
of $Imp^{x_c}$ or $Imp_s^{x_c}$ and thus highlight context-redundant
(resp. context-complementary) interactions. The darker the color, the higher
the absolute value of $Imp^{x_c}$ or $Imp_s^{x_c}$.

Respectively 49 and 26 context-dependent interactions are found in
Figures~\ref{fig:imp4}(a) and (b). In comparison, only 3 and 4 interactions are
found respectively in Figures~\ref{fig:imp4}(c) and (d) using the more
straightforward score $Imp^{x_c}$. Only 1 interaction is common between
Figures~\ref{fig:imp4}(a) and (c), while 3 interactions are common between
Figures~\ref{fig:imp4}(b) and (d). The much lower sensitivity of $Imp^{x_c}$
with respect to $Imp^{|x_c|}$ was expected given the discussions in
Section~\ref{sec:identification}. Although more straightforward, the score $Imp^{x_c}(X_m)$,
defined as the difference $Imp(X_m)-Imp(X_m|X_c=x_c)$, indeed suffers from the
fact that $Imp(X_m)$ and $Imp(X_m|X_c=x_c)$ are estimated from different
ensembles and thus do not explore the same conditionings in finite
setting. $Imp^{x_c}$ also does not have the same guarantee as $Imp^{|x_c|}$ to
find all context-dependent variables.



\section{\textsc{Conclusions}}
\label{sec:conclusions}

In this work, our first contribution is a formal framework defining and
characterizing the dependence to a context variable of the relationship between
the input variables and the output (Section~\ref{sec:contextual-relevance}). As
a second contribution, we have proposed several novel adaptations of random
forests-based variable importance scores that implement these definitions and
characterizations and we have derived performance guarantees for these scores
in asymptotic settings (Section~\ref{sec:contextual-relevance-trees}). The
relevance of these measures was illustrated on several artificial and real
datasets (Section~\ref{sec:experiments}).

There remain several limitations to our framework that we would like to address
as future works.  All theoretical derivations in Sections
\ref{sec:contextual-relevance} and \ref{sec:contextual-relevance-trees} concern
categorical input variables. It would be interesting to adapt our framework to
continuous input variables, and also, probably with more difficulty, to
continuous context variables. Finally, all theoretical derivations are based on
forests of totally randomized trees (for which we have an asymptotic
characterization). It would be interesting to also investigate non totally
randomized tree algorithms (e.g., \cite{breiman2001random}'s standard Random
Forests method) that could provide better trade-offs in finite settings.

{\footnotesize \textbf{Acknowledgements.} Antonio Sutera is a recipient of a FRIA grant from the FNRS
(Belgium) and acknowledges its financial support. This work is supported by
PASCAL2 and the IUAP DYSCO, initiated by the Belgian State, Science Policy
Office. The primary tumor data was obtained from the University Medical
Centre, Institute of Oncology, Ljubljana, Yugoslavia. Thanks go to M. Zwitter
and M. Soklic.}

\bibliographystyle{apalike}
\bibliography{bib}
\vfill

\end{document}